\documentclass{article}

\usepackage{arxiv}

\usepackage[utf8]{inputenc} 
\usepackage[T1]{fontenc}    
\usepackage{hyperref}       
\usepackage{url}            
\usepackage{booktabs}       
\usepackage{amsfonts}       
\usepackage{nicefrac}       
\usepackage{microtype}      
\usepackage{lipsum}		
\usepackage{multicol}
\usepackage{graphicx}
\usepackage[numbers,square]{natbib}
\usepackage{doi}
\usepackage{subcaption}
\usepackage{cleveref}
\usepackage{float}

\title{Finding the Subjective Truth}

\author{ 
    Dimitrios ~Christodoulou
    \\
	\texttt{dimitrios@rapidata.ai} \\
	\And
	\href{https://orcid.org/0009-0008-3406-0104}{\includegraphics[scale=0.06]{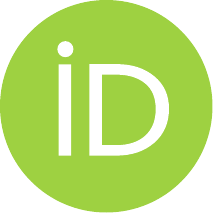}\hspace{1mm}Mads A.~Kuhlmann-Jørgensen} \\
	\texttt{mads@rapidata.ai} \\
}

\date{}

\renewcommand{\headeright}{}
\renewcommand{\undertitle}{Collecting 2 Million Votes for Comprehensive Gen-AI Model Evaluation}
\renewcommand{\shorttitle}{The Subjective Truth}

\hypersetup{
pdftitle={Finding the Subjective Truth},
pdfsubject={q-bio.NC, q-bio.QM},
pdfauthor={Dimitrios ~Christodoulou,Mads A.~Kuhlmann-Jørgensen},
pdfkeywords={Text-to-Image Models, Benchmark, Large Scale Annotation, RLHF, Human Feedback, Generative AI Evaluation},
}

\begin{document}
\maketitle

\begin{abstract}
	Efficiently evaluating the performance of text-to-image models is difficult as it inherently requires subjective judgment and human preference, making it hard to compare different models and quantify the state of the art. Leveraging Rapidata's technology, we present an efficient annotation framework that sources human feedback from a diverse, global pool of annotators. Our study collected over 2 million annotations across 4,512 images, evaluating four prominent models (DALL-E 3, Flux.1, MidJourney, and Stable Diffusion) on style preference, coherence, and text-to-image alignment. We demonstrate that our approach makes it feasible to comprehensively rank image generation models based on a vast pool of annotators and show that the diverse annotator demographics reflect the world population, significantly decreasing the risk of biases.
\end{abstract}


\begin{figure*}[h]
    \centering
    \begin{subfigure}[t]{0.245\linewidth}
        \centering
        \includegraphics[width=\linewidth]{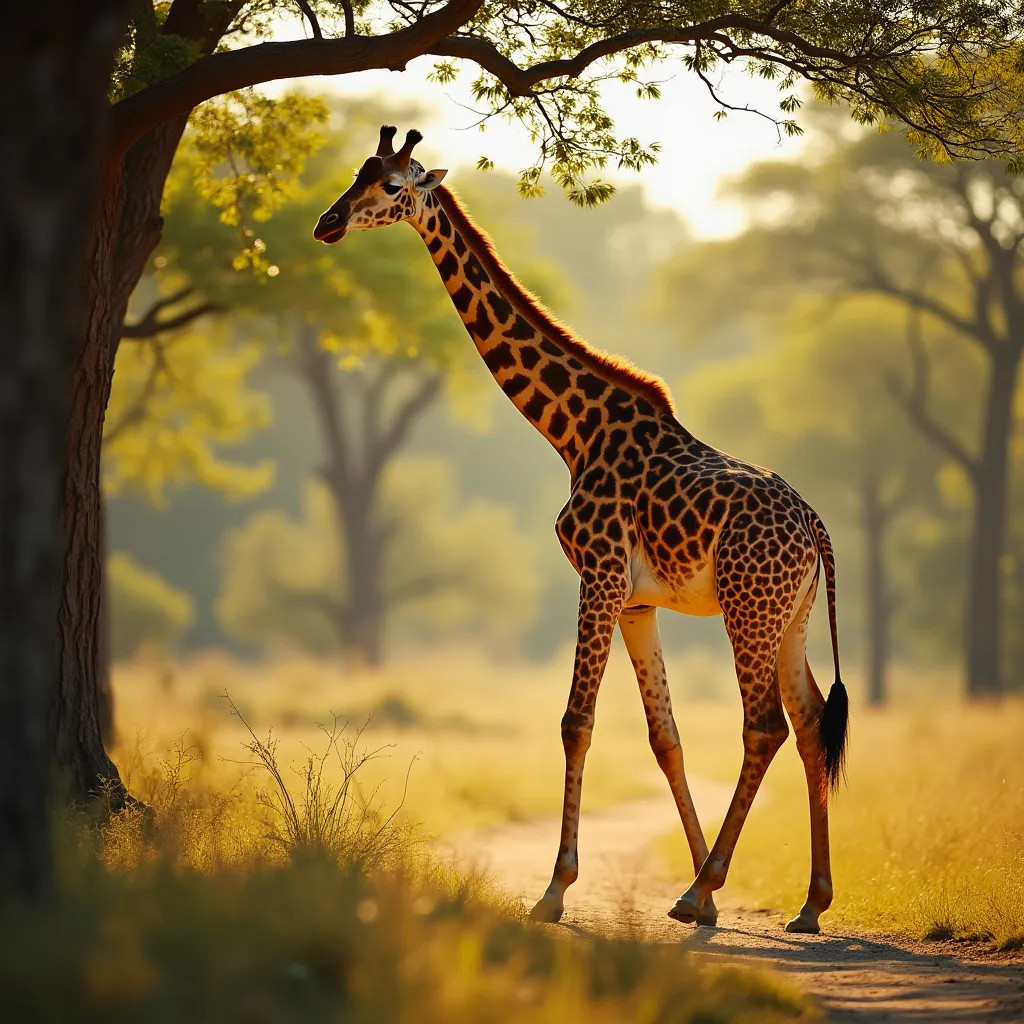}
        \caption{Flux.1}
        \label{fig:giraffe-flux}
    \end{subfigure}
    \hfill
    \begin{subfigure}[t]{0.245\linewidth}
        \centering
        \includegraphics[width=\linewidth]{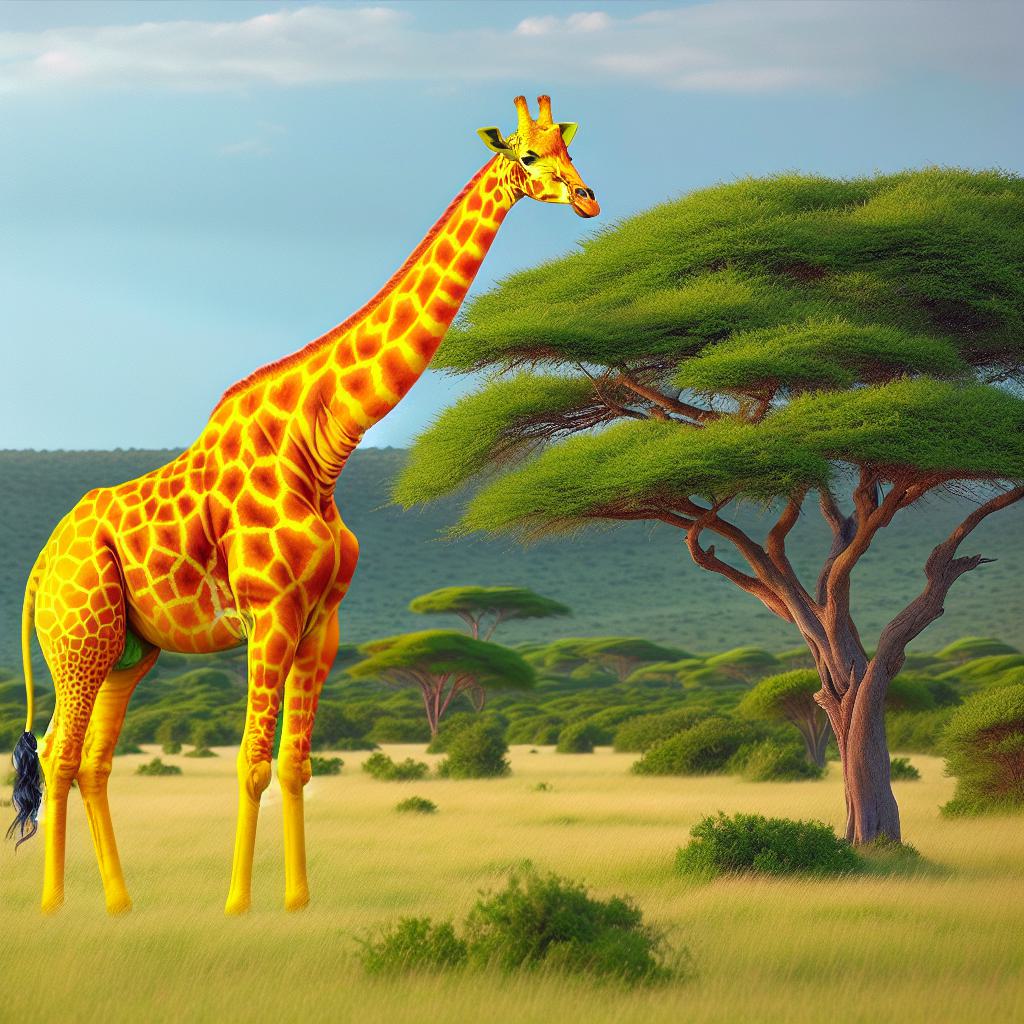}
        \caption{DALL-E 3}
        \label{fig:giraffe-dalle}
    \end{subfigure}
    \hfill
    \begin{subfigure}[t]{0.245\linewidth}
        \centering
        \includegraphics[width=\linewidth]{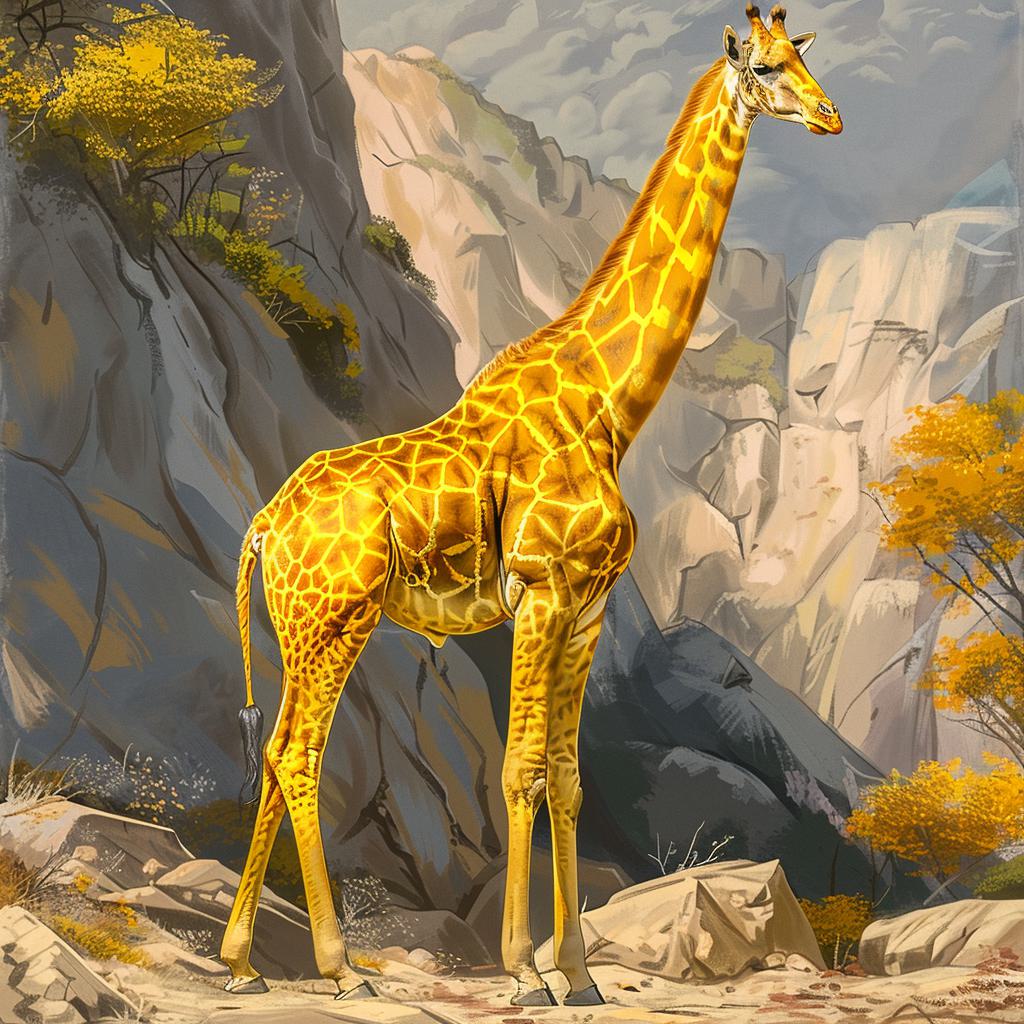}
        \caption{MidJourney}
        \label{fig:giraffe-mj}
    \end{subfigure}
    \hfill
    \begin{subfigure}[t]{0.245\linewidth}
        \centering
        \includegraphics[width=\linewidth]{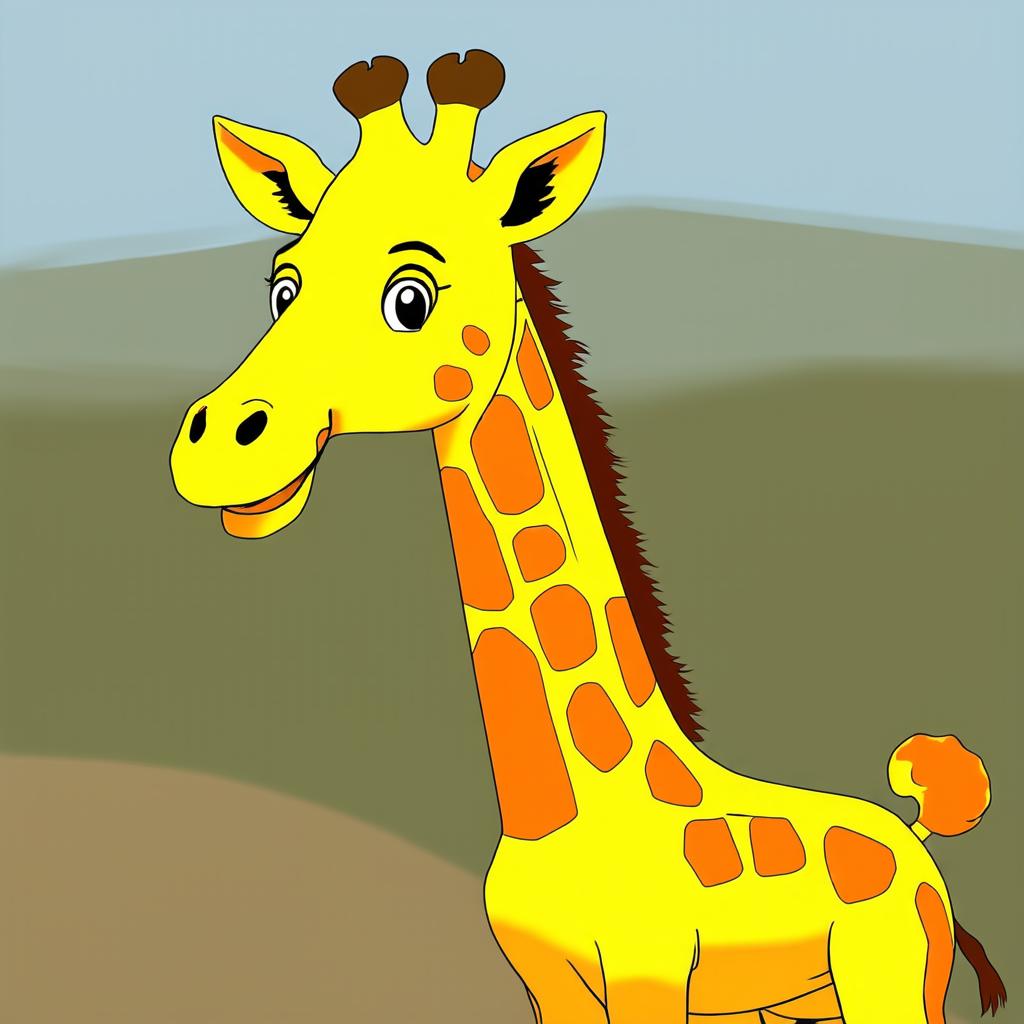}
        \caption{Stable Diffusion}
        \label{fig:giraffe-sd}
    \end{subfigure}
    \caption{Example outputs from the four evaluated models based on the prompt: \textit{A yellow colored giraffe}.}
    \label{fig:giraffe}
\end{figure*}

\begin{multicols}{2}
\section{Introduction}
Generative models are a groundbreaking class of artificial intelligence systems designed to produce new data instances based on extensive training data. Among these, text-to-image models such as DALL-E 3 \cite{dalle-3}, Stable Diffusion 3 \cite{esser2024scalingrectifiedflowtransformers}, MidJourney, and the newest, Flux.1, have gained immense popularity for their ability to create detailed images from textual descriptions. These models use deep learning to generate images from text. Each new model iteration aims to outperform the predecessor in generating high quality images. However, while the quality in a few cases can be quantified objectively, e.g. number of certain objects in the generated image, the image quality must generally be judged subjectively. This makes it non-trivial to benchmark new models against each other and ensure continuous improvement of the state of the art.

Meaningful benchmarks have played an important role in maturing other areas of deep learning research such as computer vision where the community has settled on a few standardized benchmarks such as ImageNet for image classification and COCO for object detection. A similar standardized benchmark does not exist for text-to-image generative models. Instead, authors often propose their own evaluation setup when publishing the results of their newest models \cite{dalle-3}. This makes it difficult to gain an objective overview of the current state of the art. However, the subjective nature of the evaluation makes it difficult to introduce a standardized large scale benchmark. 
The main solutions in literature include soliciting human opinions from a user panel, or using AI evaluation models trained to predict human preferences \cite{dalle-3, lin2023designbenchexploringbenchmarkingdalle, nichol2022glidephotorealisticimagegeneration, rombach2022highresolutionimagesynthesislatent}. However, both methods struggle to generalize well and capture the full spectrum of human preferences and cultural backgrounds. Human-based evaluations often lack sufficient demographic information and due time and budget constraints are limited to a small number of participants, making them susceptible to biases and potential stereotypes. On the other hand, AI evaluation models naturally lack human understanding, can perpetuate biases, and may not fully capture contextual and nuanced quality assessments. By design, these models are trained on last generations’ data, and thus are not guaranteed to reflect human opinions on new image generation models. In short, existing evaluation methods are fragmented and not guaranteed to accurately measure the true metric of broad human preferences within necessary time and cost constraints.

With Rapidata’s technology \cite{Rapidata}, we introduce a new paradigm in evaluating text-to-image generative models. By accessing a vast, diverse pool of annotators from around the world, we provide access to human feedback at previously unseen scale and speed from a wide array of cultural backgrounds. With this, we present what to the best of our knowledge is the first, large scale, truly representative benchmark for text-to-image generation that can feasibly be repeated continuously with introduction of new models. We provide the following contributions:
\begin{itemize}
    \item Introduction of a novel annotation process for large scale human preference collection at high speed and low cost.
    \item A carefully curated set of 282 image generation prompts assembled from existing literature to cover a wide range of important evaluation criteria.
    \item A ranking of the most prevalent text-to-image generative models, Flux.1, DALL-E 3, Stable Diffusion, and MidJourney based on 2 million human votes collected for 4512 generated images.
\end{itemize}

With over 2 million votes collected, this is the most comprehensive text-to-image generation benchmark by orders of magnitude, with most existing publications sporting numbers in the 2-20k range \cite{dalle-3,lin2023designbenchexploringbenchmarkingdalle}.

\section{Methodology}
\label{sec:method}

\subsection{Annotation Framework}
The benchmark is designed to evaluate the quality of text-to-image outputs over a range of meaningful settings based on three criteria common in literature \cite{10431766, dalle-3} – Style, Coherence, and Text-to-Image alignment. Concretely, this is done by repeatedly presenting annotators with two  generated images and asking them to select the best option based on one of the following questions respectively:
\begin{enumerate}
    \item Which image do you prefer?
    \item Which image is more plausible to exist and has fewer odd or impossible-looking things?
    \item Which image better reflects the caption above them?
\end{enumerate}

\subsubsection{Interface Design}
To ensure quick and efficient data collection, we have designed the annotation interface to be intuitive and accessible to users with little or no introduction. Inspired by existing works, we decide to always have annotators make a choice between two presented images, as this is deemed more intuitive than e.g. rating individual images. Thus, the base design consists of a criteria question at the top, with the two image options displayed below. In addition, some variations were needed to e.g. display the prompt for text-to-image alignment and ensure higher data quality. 
\begin{enumerate}
    \item \textbf{Base Design}: used for the Preference and Coherence criteria: Participants are shown pairs of images generated by different models and asked to select the one that better fits the criteria. They are not provided with the prompt used for generating the image as this is not relevant for style preference and coherence.
    \item \textbf{Display Prompt}: used for the Text-to-Image Alignment criteria: In this task, users are given a prompt and must choose which image best matches the given description. The core structure remains the same, but with an additional feature: an animated prompt displayed word-by-word at the top of the screen. This dynamic presentation of the prompt engages users and ensures that they fully comprehend the description before making a selection between the two images.
    \item \textbf{Validation Tasks}: These tasks are embedded to test the attentiveness and engagement of users to ensure higher quality responses. They are crafted to be simple, where the correct choice is obvious, and the annotator will be punished for a wrong answer, ensuring they are actively participating in the task.
\end{enumerate}

\begin{figure*}
    \centering
    \begin{subfigure}[t]{0.32\linewidth}
        \centering
        \includegraphics[width=\linewidth]{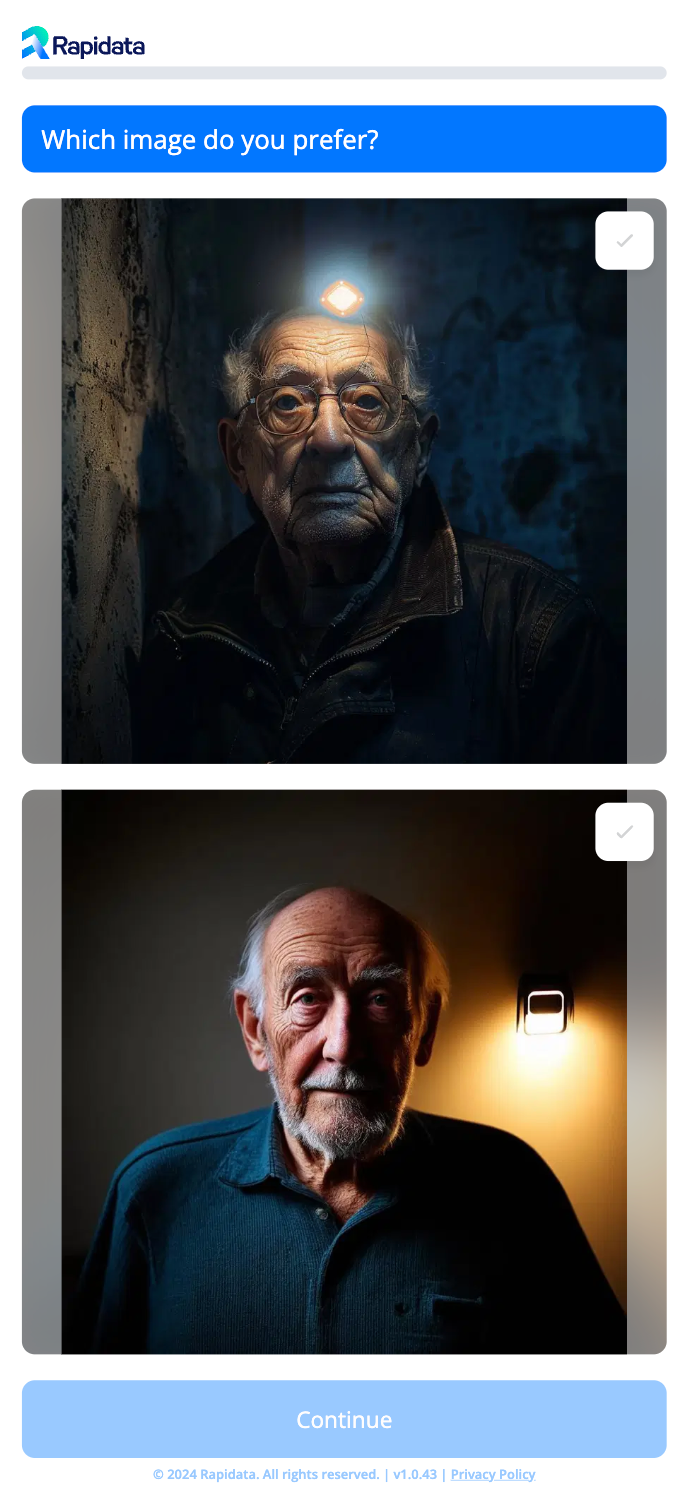}
        \caption{\textbf{Preference}: As criteria, the annotators are asked the question: '\textit{Which image do you prefer?}' displayed at the top. Below are the options; images generated by two different models. After selecting their preferred image, users can click 'Continue' to proceed.}
        \label{fig:preference}
    \end{subfigure}
    \hfill
    \begin{subfigure}[t]{0.32\linewidth}
        \centering
        \includegraphics[width=\linewidth]{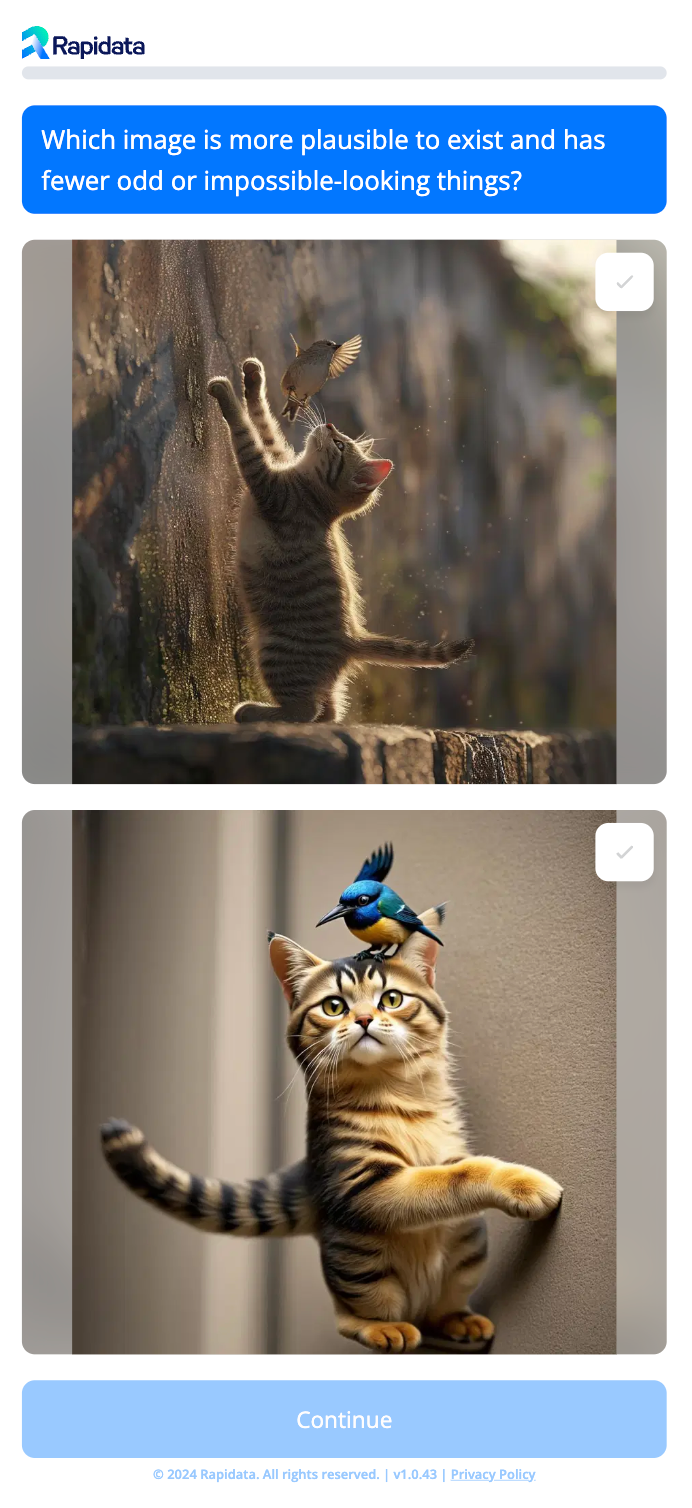}
        \caption{\textbf{Coherence}: Similar to the preference task, but with a different criteria question displayed at the top; ‘\textit{Which image is more plausible to exist and has fewer odd or impossible-looking things?}'}
        \label{fig:coherence}
    \end{subfigure}
    \hfill
    \begin{subfigure}[t]{0.32\linewidth}
        \centering
        \includegraphics[width=\linewidth]{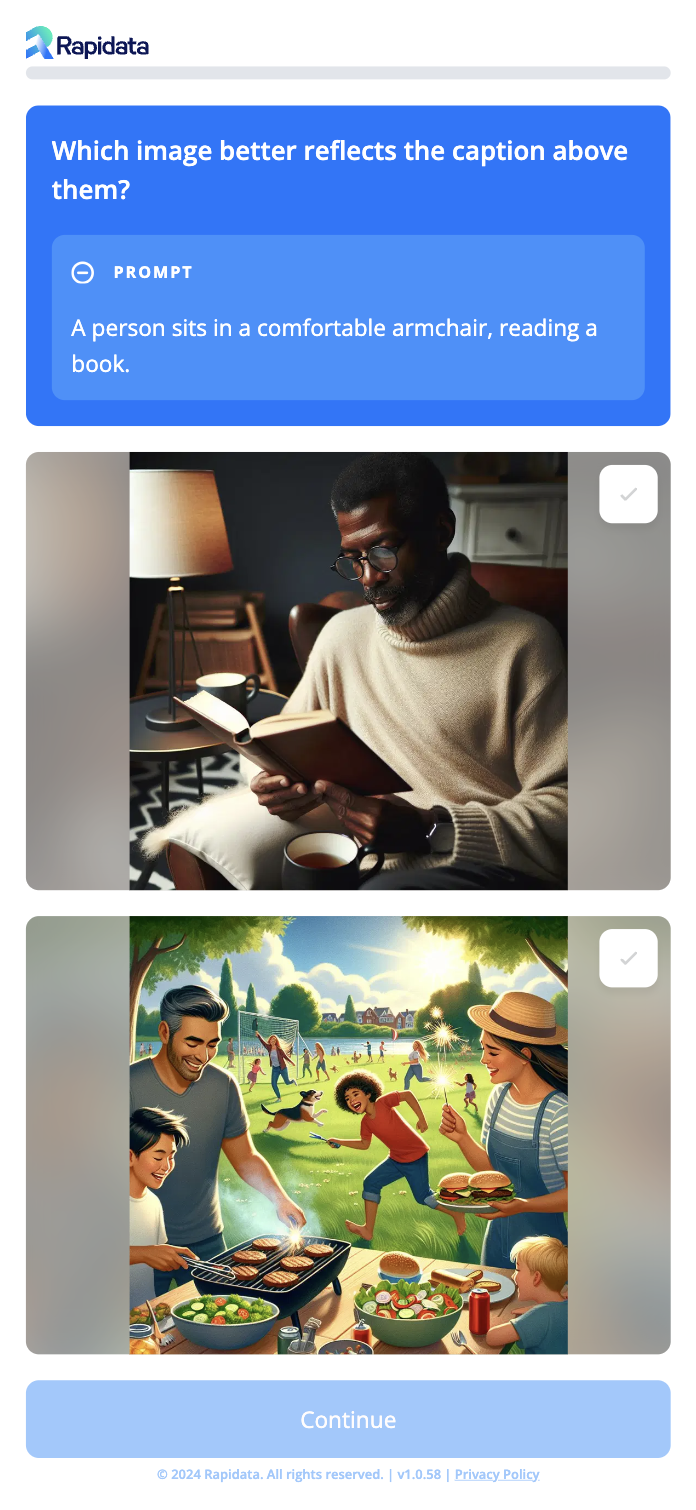}
        \caption{\textbf{Text-to-Image Alignment}:The task has the criteria question: '\textit{Which image better reflects the caption above them?}' The prompt is shown for context and appears quickly word-by-word to grab attention. Users click 'Done' after reading to indicate readiness. Then, images become selectable.}
        \label{fig:text_alignment}
    \end{subfigure}
    \caption{Illustration of the interface used for annotation.}
    \label{fig:annotation-interface}
\end{figure*}

\Cref{fig:annotation-interface} provides a detailed explanation of the task design and methodology used for the different criteria in our annotation process.

To improve efficiency, we group multiple comparisons into a single session. A session consists of a set of tasks assigned to an annotator in one sitting. Depending on difficulty,  each session can include up to three tasks – all with the same criteria. For instance, participants may complete up to three preference or coherence tasks, or two text-to-image alignment tasks (with the first being a validation task to ensure engagement). On average, the preference or coherence sessions take 12.5 seconds, while the two text-to-image alignment tasks take about 15 seconds.

\subsection{Distribution}
Our annotation process leverages crowdsourcing techniques to gather detailed assessments of text-to-image model performance from a broad and diverse user base. By distributing tasks through mobile apps, we ensure accessibility and ease of participation, allowing us to tap into a global pool of annotators across different demographics, geographic regions, and cultural backgrounds. This approach not only broadens the diversity of perspectives in the evaluations but also increases scalability, enabling us to handle large volumes of data in a cost-effective manner.

\subsection{Quality Assurance}
To enhance the reliability of the data collected and mitigate the risk of contributions from adversarial users, we implement several quality control measures.

\textbf{Time-Based Controls}\\
A minimum expected time is set for completing each task. If a participant submits a response faster than this threshold, a mandatory five-second penalty is enforced before allowing them to proceed. This ensures that users dedicate adequate time to consider each image pair or prompt, discouraging quick and thoughtless responses.

\textbf{Validation Questions}\\
To confirm that users are genuinely engaged, we incorporate validation tasks that are simple and intentionally straightforward, with an obvious correct answer. Annotators who do not successfully complete these tasks may be flagged for further review or disqualified from continuing to contribute

\textbf{Global Reach}\\
To accommodate our diverse pool of global annotators, all tasks are presented in the user’s preferred language, typically based on the default language setting of their mobile device. To ensure accessibility, all prompts are translated into the user's device language. This ensures that annotators can fully understand and engage with the tasks, regardless of their language background.

\subsection{Prompts}
The benchmark is organized into distinct categories, each tailored to evaluate critical attributes for assessing generative AI models. We base the input prompts on the existing benchmark, \textit{DrawBench} \cite{saharia2022photorealistictexttoimagediffusionmodels}, however, to ensure a comprehensive evaluation, we enhanced it by integrating prompts from various complementary benchmarks. This allowed for a more comprehensive and diverse assessment of model capabilities across a wide range of attributes. While DrawBench covers key aspects such as \textbf{color accuracy, handling of descriptions, misspellings, counting, rare words}, and \textbf{conflicting scenarios}, it lacks coverage for certain nuanced attributes. To address this, we incorporated additional datasets, as guided by ImagenHub \cite{ku2024imagenhubstandardizingevaluationconditional}, ensuring a broader evaluation of model capabilities: 

\textbf{DiffusionDB} \cite{wang2023diffusiondblargescalepromptgallery}: Introduces a variety of \textbf{general descriptions}, enhancing model testing across a wide range of scenarios. Example: “\textit{average man mugshot photo.}”
    
\textbf{ABC-6K} \cite{feng2023trainingfreestructureddiffusionguidance}: Provides hard prompts with \textbf{complex details}, challenging the model’s ability to manage intricate descriptions. Example: “\textit{Male tennis player with white shirt and blue shorts swinging a black tennis racket.}”

\textbf{HRS-Bench}\cite{bakr2023hrsbenchholisticreliablescalable}: Focuses on evaluating \textbf{fidelity, emotion, bias, size, creativity}, and \textbf{counting}, crucial for assessing more subjective and contextual aspects of image generation. Example: “\textit{A real scene about a laptop, which makes us feel anger.}”

\textbf{T2I-CompBench} \cite{huang2023t2icompbenchcomprehensivebenchmarkopenworld}: Tests key structural elements such as\textbf{ color, actions, shapes, texture}, and \textbf{spatial/non-spatial relationships}, challenging models with both abstract and realistic prompts. Example: “\textit{The fluffy pillow was on top of the hard chair.}”

\textbf{DALLE3-EVAL} \cite{dalle-3-eval}: Focuses on \textbf{complex} and varied \textbf{descriptions of celebrities}. Example: “\textit{Adele, wearing a chef's apron, experiments with exotic ingredients in a modern kitchen. The vibrant colors and textures of the food contrast with her focused expression as she creates a culinary masterpiece.}”.

\subsection{Image Generation}
For image generation we applied the same text prompts across all models. We used APIs for DALL-E 3, MidJourney, and Flux.1. For Stable Diffusion we used the model through Hugging Face. This standardized method allowed for direct comparison of each model’s ability to interpret and visually render the prompts.
\begin{itemize}
    \item \textbf{Flux.1}: Used via the Replicate API with the 'flux-pro' model, standard image quality, and dimensions set to 1024x1024 pixels.
    \item \textbf{DALL-E 3}: Accessed via OpenAI's API with 'dall-e-3', standard image quality, and 1024x1024 pixel dimensions. We generated multiple images per prompt through parallel requests.
    \item \textbf{MidJourney}: Employed through the ImaginePro API with 'MidJourney Version 5.2', an aspect ratio of 1:1, minimal chaos, quality set to 1, and stylization at 100. Four upscaled images were selected for comparison.
    \item \textbf{Stable Diffusion}: Used via Stability AI on Hugging Face with 'stable-diffusion-3-medium-diffusers' and a guidance scale of 7, producing 1024x1024 pixel images, dropping the T5 Text Encoder during Inference.
\end{itemize}

\section{Results\protect\footnote{NOTE: After finalizing the experiments and paper, an error has been discovered which resulted in 19 of the prompts for Stable Diffusion to be corrupted, resulting in bland or senseless generated images. In total, this affects about $1.5$\% of the comparisons. While this is extremely unfortunate we believe that the effect on the overall results is negligible, and the general takeaways remain the same.}}
\subsection{Data}
To provide a comprehensive comparison, we generated four images for each model and prompt. For each prompt, each image was paired against the 12 images from the other models, resulting in 96 pairwise comparisons per prompt for each criteria. Across the entire dataset of 282 prompts, this led to 27,072 total comparisons per criteria.

For each comparison, we collected 26 votes, leading to 2,496 votes per prompt. With 282 prompts in total, we accumulated over 700,000 votes per criteria, resulting in more than 2 million votes across the three criterias. Our research involved a vast and diverse group of 144,292 participants from 145 different countries, reflecting a truly global representation.

\subsection{Ranking Algorithm}
To derive a ranking for the models based on the collected votes, we employed the Iterative Bradley-Terry ranking algorithm \cite{JMLR:v24:22-1086}. This algorithm is well-suited for scenarios where data is collected in full before determining the ranking, whereas e.g. Elo is a typical choice for online use cases. 
Using the Bradley-Terry model we assign probabilistic scores to each image generation model. Based on these scores, the probability that model $i$ beats model $j$ in a pairwise comparison can be found as $P(i>j)= \frac{p_i}{p_i+p_j}$, where $p_i$ and $p_j$ are the scores assigned to model $i$ and $j$ respectively. There is no known closed form solution for calculating the score from a set of pairwise comparisons, however, the problem can be solved by iteration. At each iteration, the score for each model is updated according to \cref{eq:bradley-terry}

\begin{equation}
\label{eq:bradley-terry}
    p'_i = \frac{\sum_{j \neq i} w_{ji} \frac{p_j}{p_i + p_j}}{\sum_{j \neq i} w_{ji} \frac{1}{p_i + p_j}},
\end{equation}

where $w_{ij}$ is the number of comparisons where model $i$ beat model $j$. Since the scores are arbitrary up to a common factor, we decide to normalize the scores to sum to 100.

\begin{table*}
  \centering
  \begin{tabular}{lcccc}
    \toprule
    Criteria \textbackslash{} Model & Flux.1 & DALL-E 3 & MidJourney & Stable Diffusion \\
    \midrule
    Preference & \textbf{29.86} & 24.17 & 23.98 & 21.99 \\
    Coherence & \textbf{29.61} & 22.92 & 23.37 & 24.09 \\
    Text-Image Alignment & \textbf{27.36} & 26.76 & 24.48 & 21.40 \\
    \bottomrule
  \end{tabular}
  \caption{The Bradley-Terry score for each of the four image generation models across the three different criteria.}
  \label{tab:model-comparison}
\end{table*}

\subsection{Ranking}
\begin{figure}[H]
    \centering
    \includegraphics[width=1\linewidth]{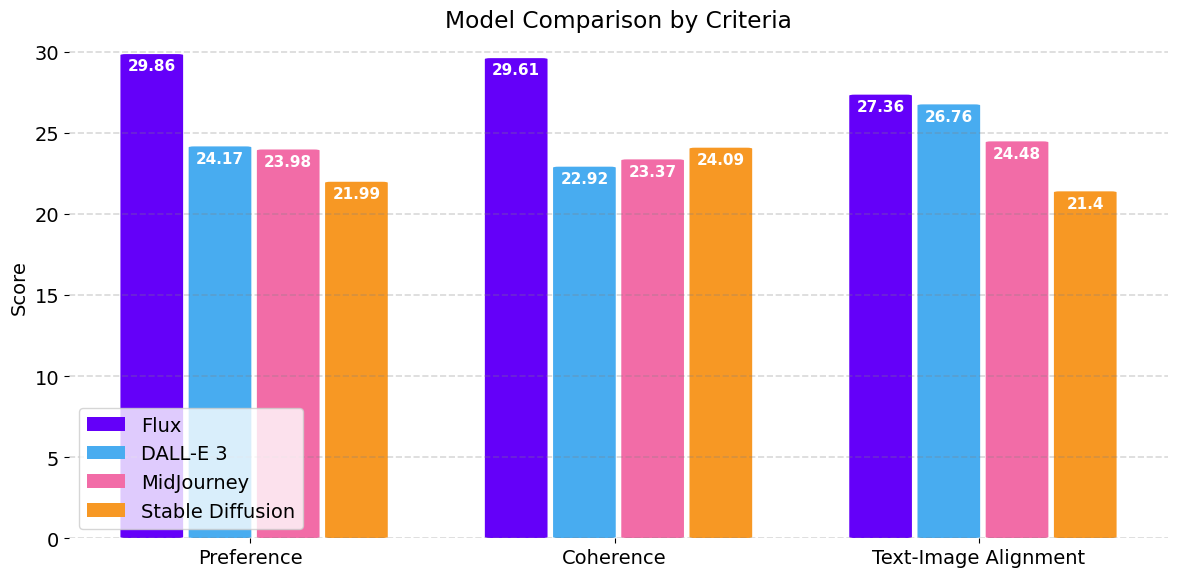}
    \caption{Ranking of the four models, Flux.1, DALL-E 3, MidJourney, and Stable Diffusion, based on the three criteria, style, coherence, and text-to-image alignment.}
    \label{fig:ranking-graph}
\end{figure}

Using the Bradley-Terry ranking method, we derived the final rankings of these models. The summary of the results is presented in \cref{tab:model-comparison,fig:ranking-graph}, highlighting the relative performance of each model across the three criteria.

\begin{figure*}
    \centering
    \begin{subfigure}[t]{0.245\linewidth}
        \centering
        \includegraphics[width=\linewidth]{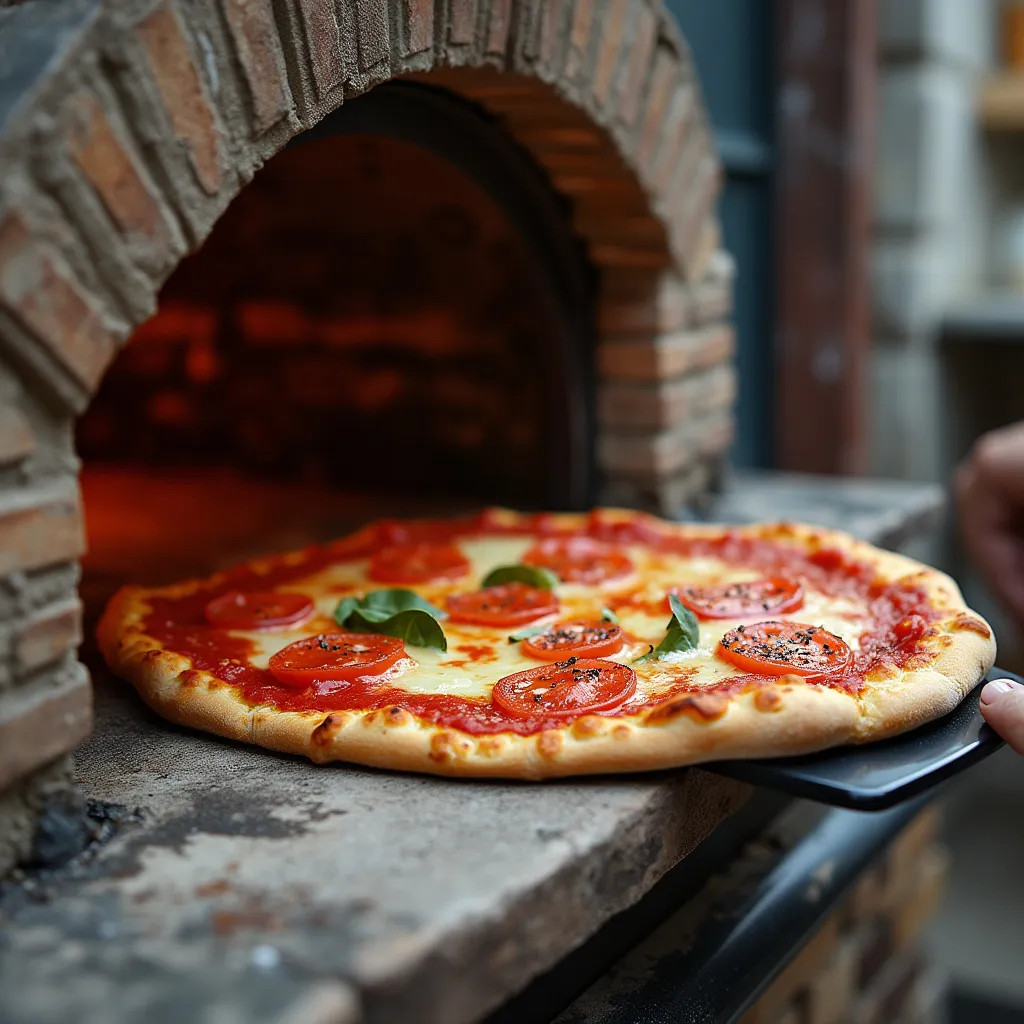}
        \caption{Flux.1}
        \label{fig:pizza-flux}
    \end{subfigure}
    \hfill
    \begin{subfigure}[t]{0.245\linewidth}
        \centering
        \includegraphics[width=\linewidth]{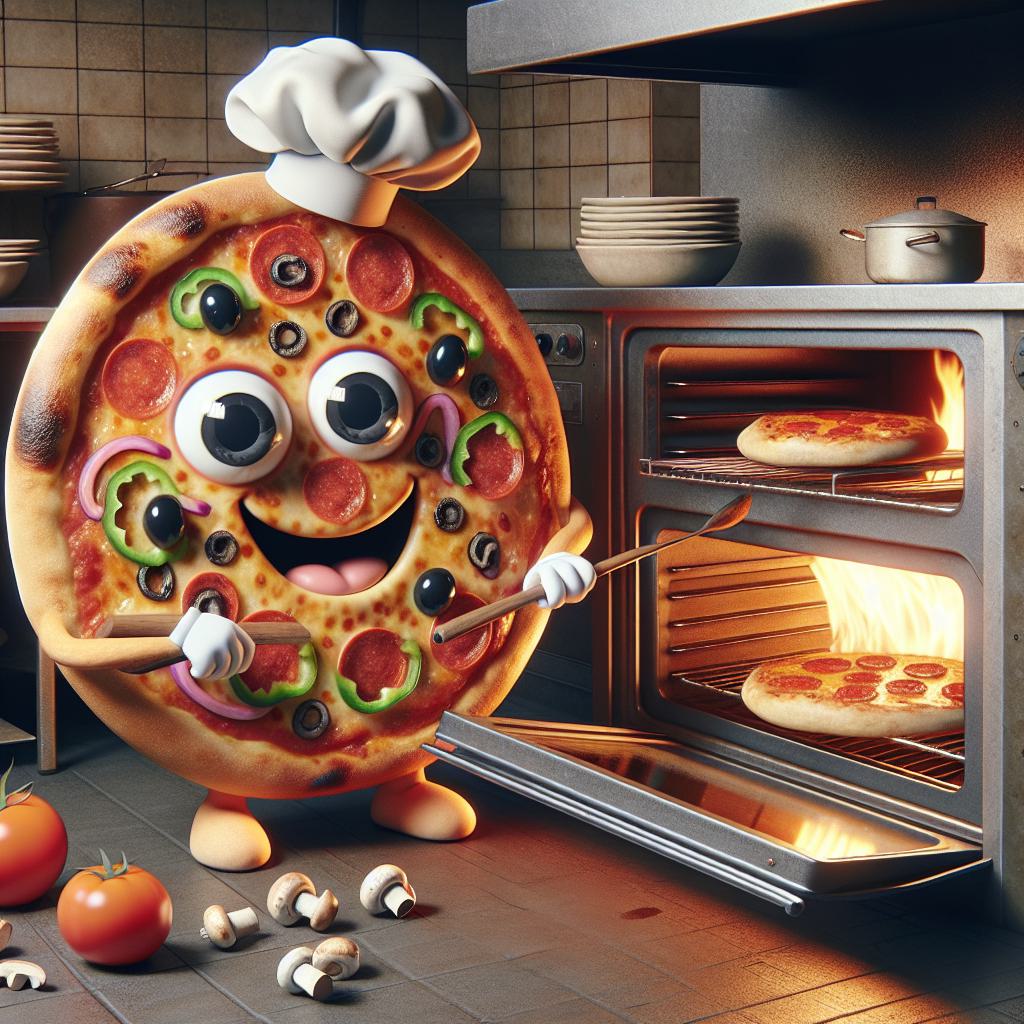}
        \caption{DALL-E 3}
        \label{fig:pizza-dalle}
    \end{subfigure}
    \hfill
    \begin{subfigure}[t]{0.245\linewidth}
        \centering
        \includegraphics[width=\linewidth]{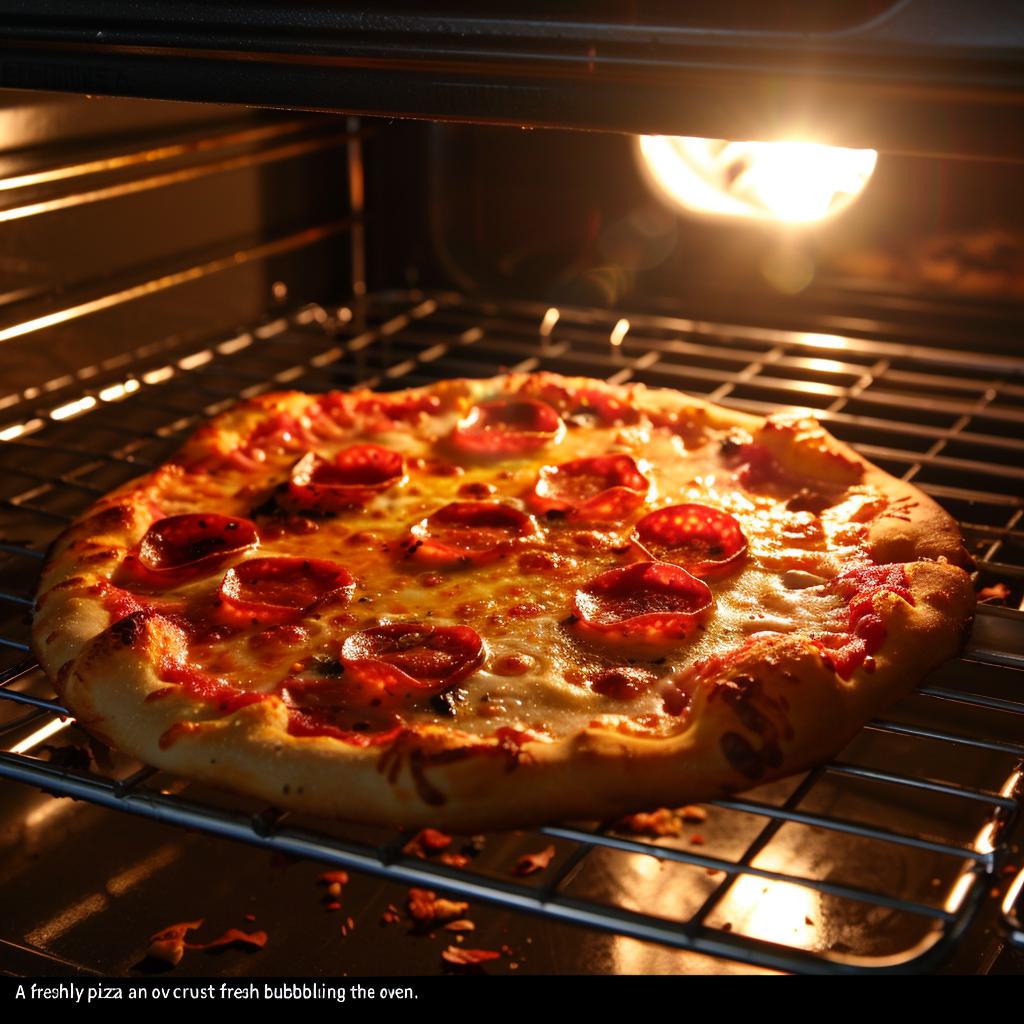}
        \caption{MidJourney}
        \label{fig:pizza-mj}
    \end{subfigure}
    \hfill
    \begin{subfigure}[t]{0.245\linewidth}
        \centering
        \includegraphics[width=\linewidth]{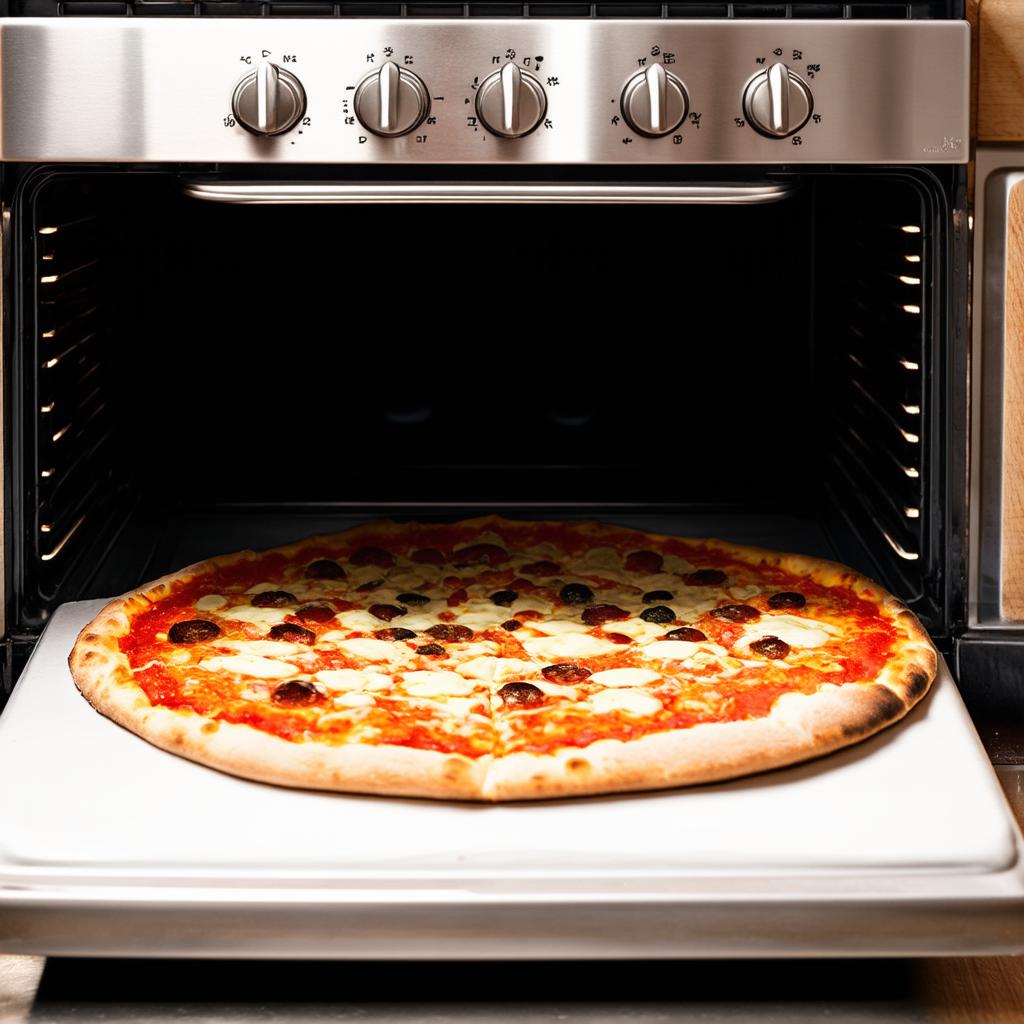}
        \caption{Stable Diffusion}
        \label{fig:pizza-sd}
    \end{subfigure}
    \caption{Example outputs from the four evaluated models based on the prompt: \textit{A pizza cooking an oven}}
    \label{fig:pizza}
\end{figure*}

\textbf{Preference}: The newest model, Flux.1, significantly outperformed the other three in terms of style preference, achieving the highest ranking with a score of 29.86. The Stable Diffusion model ranked considerably lower than the others at a score of 21.99, whereas DALL-E 3 and MidJourney placed very close. While style preference is obviously subjective, these results are also reflected particularly in \cref{fig:giraffe,fig:pizza}

\textbf{Coherence}: In terms of image coherence, the Flux.1 model again outperformed the others with a score of 29.61. Interestingly however, Stable Diffusion scored higher than both DALL-E 3 and MidJourney for this criteria at 24.09, indicating that while performing the worst in the other two criterias, Stable Diffusion may generally be less likely to generate incoherent and implausible artifacts than DALL-E 3 and MidJourney. 

\begin{figure}[H]
    \centering
    \begin{subfigure}[t]{0.475\linewidth}
        \centering
        \includegraphics[width=\linewidth]{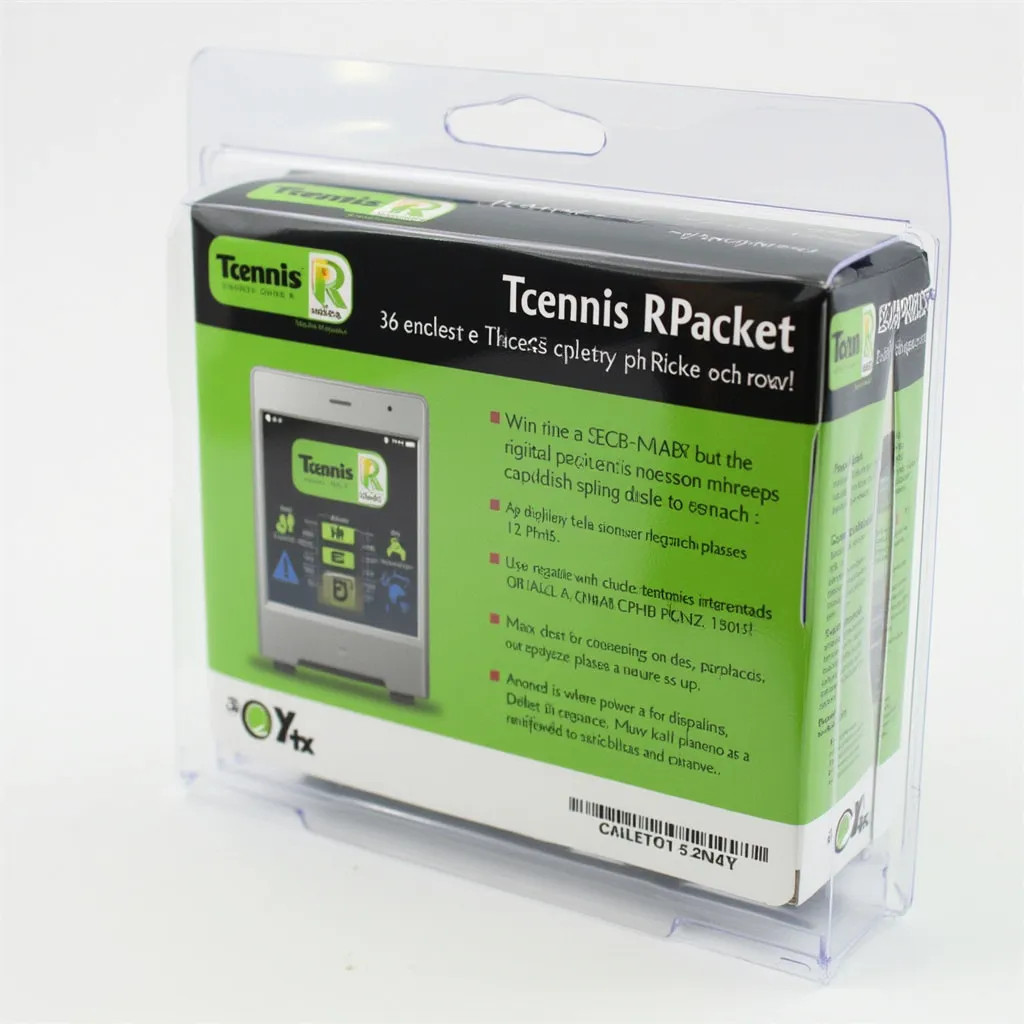}
        \caption{Flux.1}
        \label{fig:racket-flux}
    \end{subfigure}
    \hfill
    \begin{subfigure}[t]{0.475\linewidth}
        \centering
        \includegraphics[width=\linewidth]{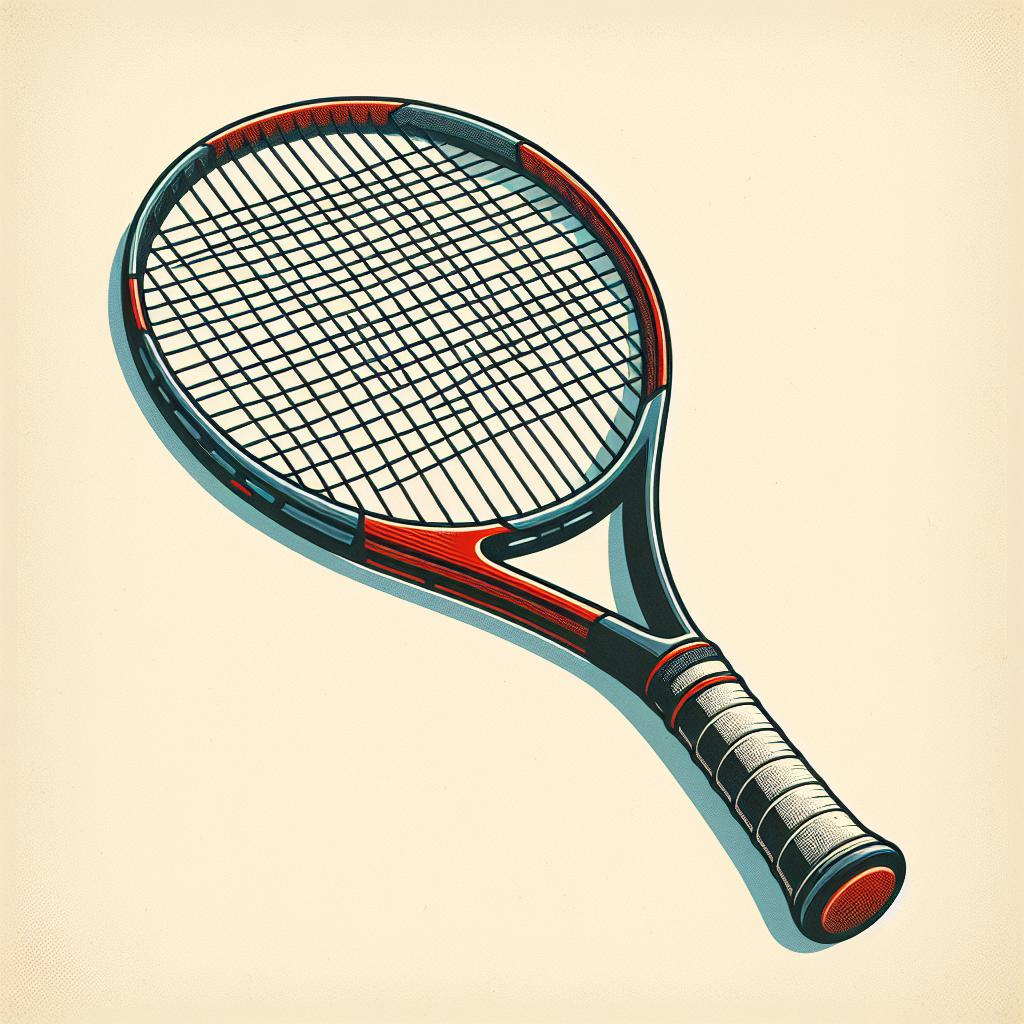}
        \caption{DALL-E 3}
        \label{fig:racket-dalle}
    \end{subfigure}
    \caption{Example outputs from Flux.1 and DALL-E 3 based on the prompt: \textit{'Tcennis rpacket'.}}
    \label{fig:racket}
\end{figure}

\textbf{Text-to-Image Alignment}: The Flux.1 model comes out on top again in text-to-image alignment, with a score of 27.36. However, the gap is not as large as for the other criterias with DALL-E 3 closely following with a score of 26.76. MidJourney and Stable Diffusion lagged behind with scores of 24.48 and 21.4, respectively. 
\Cref{fig:racket,fig:pizza}, illustrates what could be part of the reason for DALL-E 3 to bridge the gap with Flux.1 in text-to-image alignment. In cases of typos and odd sentences, from our point of view it appears that DALL-E 3 in these cases interprets the prompts more 'intuitively'. In \cref{fig:racket} DALL-E 3 interprets the misspelled prompt as '\textit{Tennis racket}' whereas Flux.1 seemingly focuses on 'packet' part of '\textit{Tcennis r\textbf{packet}}'. In \cref{fig:pizza} most models ignore the fact that the pizza is supposed to be the subject doing the cooking, however DALL-E 3 captures that aspect. It appears as if the other models instead interpreted the prompt as '\textit{A pizza cooking \textbf{in} an oven}' instead of interpreting it literally. 

Based on the results we can highlight some strengths and weaknesses among the evaluated models:
\begin{itemize}
    \item \textbf{Flux.1}: Top performer across all criterias, and by a significant margin in both style preference and coherence. This firmly cements that this should be the default choice for most applications. 
    \item \textbf{DALL-E 3}: The best performance was in terms of text-to-image alignment, although still trailing behind Flux.1. It was however the worst model in terms of producing coherent images. While there is currently little reason to choose DALL-E 3 over Flux.1, if the main focus is text-to-image alignment for odd and abstract prompts, it might in some cases be an equal or better choice.
    \item \textbf{MidJourney}: This model does not excel at any of the criterias but also is never the worst.
    \item \textbf{Stable Diffusion}: Shows relatively lower performance in all evaluated criterias, however, it remains competitive in image coherence with a score of 24.09, suggesting that it is better suited for tasks prioritizing internal consistency over user preference or text alignment.
\end{itemize}

\subsection{Demographic Statistics}
With our proposed benchmark and annotation framework, we aim to improve global representation to reduce biases. In this section we therefore present and analyze demographic data of the participating annotators. We present three statistics; geographic location, age group, and gender. Location is directly provided through the distribution method, whereas gender and age are self-reported by the annotators.

\begin{figure*}
    \centering
    \includegraphics[width=0.85\linewidth]{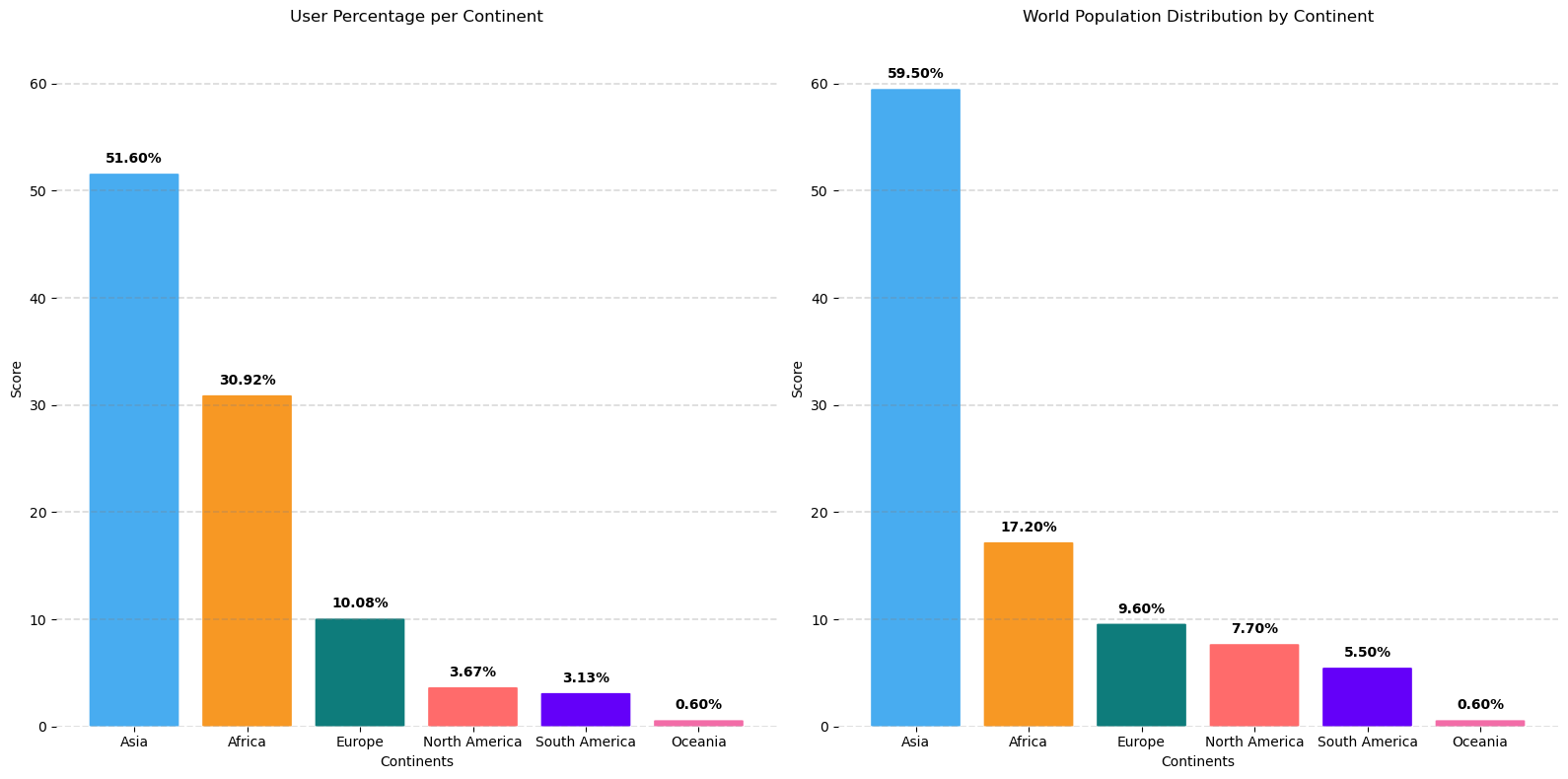}
    \caption{Overview of the distribution of annotators by continent (left) compared to the world population distribution (right)}
    \label{fig:annotator-distribution}
\end{figure*}

\textbf{Location}: For simplicity and visualization, in \cref{fig:annotator-distribution} we break down location by continent. We recognize that particularly in vast and diverse continents like Asia, this may leave out some nuances, but it provides comprehensible insights. By comparing our distribution to the world population distribution, we see that the general trend is captured, however we have an overrepresentation of African annotators and a slight underrepresentation of annotators from Asia. This is something that can be considered and adjusted in the future. We also see an underrepresentation of North America, however one could argue that this region is culturally very close to Europe, which is slightly overrepresented. Overall, the distribution represents a reasonable estimate to world population and underlines the argument that our proposed framework reaches a diverse set of annotators to reduce biases.  Additionally, for future work, we can adjust this, as our tool allows us to target specific countries or even languages, offering flexibility in tailoring tasks to particular regions or linguistic groups.

\textbf{Age}: The graph provides a breakdown of the percentage of users across various age groups who submitted votes in our evaluation. We observe a higher representation of younger annotators, which is probably to be expected from the crowd-sourcing method. Overall however, the distribution is quite even.
\begin{figure}[H]
    \centering
    \centering
    \includegraphics[width=0.9\linewidth]{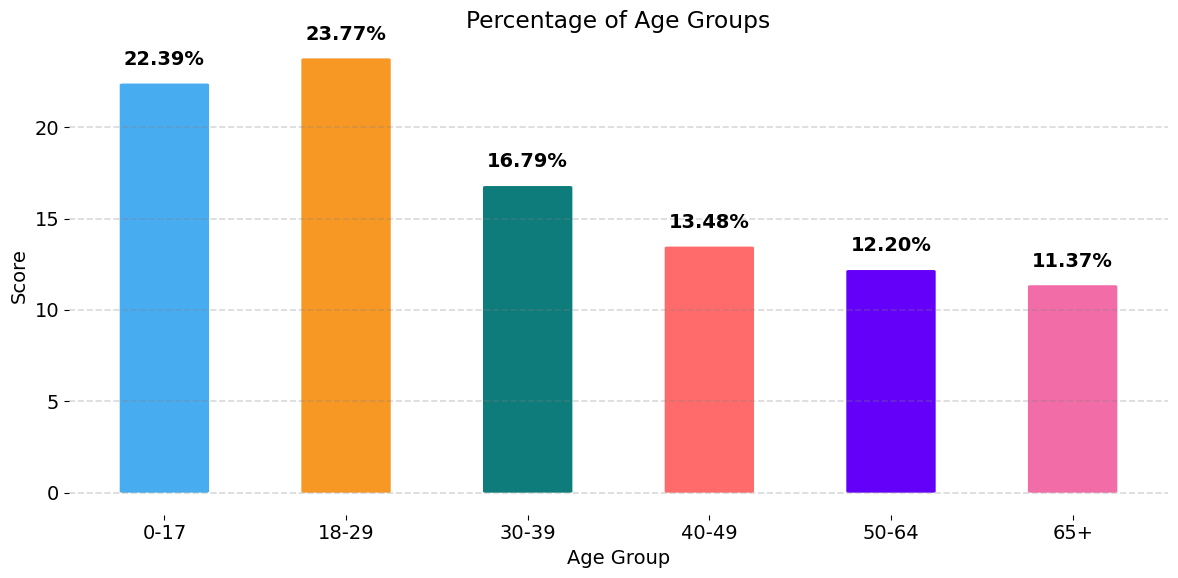}
    \caption{Distribution of the self-reported age ranges of the annotators involved.}
    \label{fig:age}
\end{figure}

\textbf{Gender}: The graph illustrates the gender distribution of users who participated in the evaluation. We observe a very even distribution between men and women. However, we also observe a large portion that answered another category or did not provide an answer. This can be taken into consideration to structure and phrase this question better in the future.
\begin{figure}[H]
    \centering
    \centering
    \includegraphics[width=0.9\linewidth]{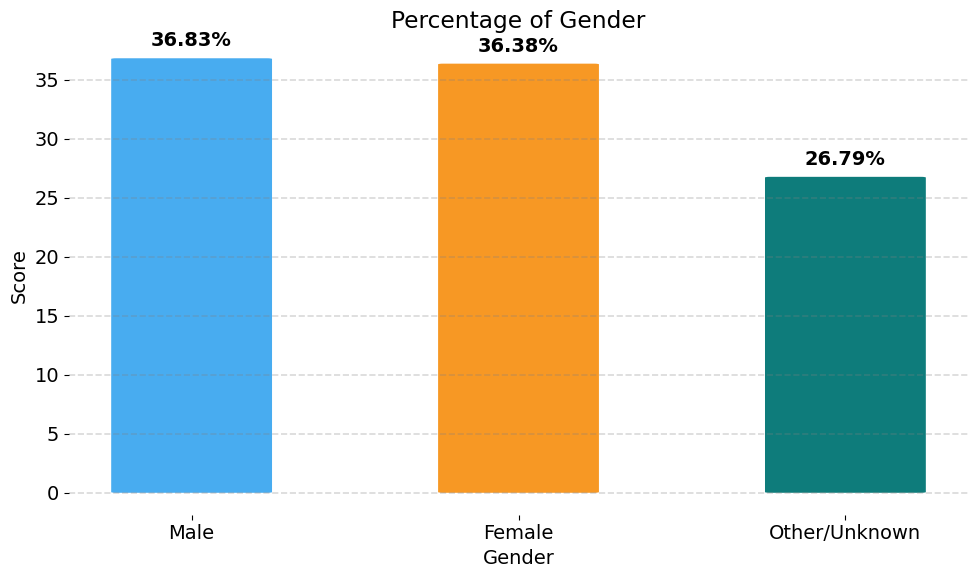}
    \caption{Distribution of the self-reported gender of the annotators involved. }
    \label{fig:gender}
\end{figure}

\section{Conclusion}
We have shown that our annotation framework and proposed benchmark provides an efficient, representative, and scalable way to comprehensively evaluate image generation models. This is only possible through the novel crowd-sourcing approach made accessible through Rapidata’s technology. In the scope of just a few days, we cost-effectively collected more than 2 million responses – to the best of our knowledge, the largest image generation model evaluation by a significant margin. 
The scale and efficiency illustrated in this work also opens the door for incorporating human feedback directly into the training process. Reinforcement learning from human feedback (RLHF) has been a significant driver in bringing LLMs to the level we see today \cite{ouyang2022traininglanguagemodelsfollow}. Rapidata’s technology provides easy access to diverse, efficient, and scalable feedback to include humans in the training loop.

\bibliographystyle{unsrtnat}
\bibliography{subjective_truth}  

\begin{thebibliography}{16}
\providecommand{\natexlab}[1]{#1}
\providecommand{\url}[1]{\texttt{#1}}
\expandafter\ifx\csname urlstyle\endcsname\relax
  \providecommand{\doi}[1]{doi: #1}\else
  \providecommand{\doi}{doi: \begingroup \urlstyle{rm}\Url}\fi

\bibitem[Betker et~al.(2023)Betker, Goh, Jing, Brooks, Wang, Li, Ouyang, Zhuang, Lee, Guo, Manassra, Dhariwal, Chu, Jiao, and Ramesh]{dalle-3}
James Betker, Gabriel Goh, Li~Jing, Tim Brooks, Jianfeng Wang, Linjie Li, Long Ouyang, Juntang Zhuang, Joyce Lee, Yufei Guo, Wesam Manassra, Prafulla Dhariwal, Casey Chu, Yunxin Jiao, and Aditya Ramesh.
\newblock Improving image generation with better captions.
\newblock 2023.
\newblock URL \url{https://cdn.openai.com/papers/dall-e-3.pdf}.

\bibitem[Esser et~al.(2024)Esser, Kulal, Blattmann, Entezari, Müller, Saini, Levi, Lorenz, Sauer, Boesel, Podell, Dockhorn, English, Lacey, Goodwin, Marek, and Rombach]{esser2024scalingrectifiedflowtransformers}
Patrick Esser, Sumith Kulal, Andreas Blattmann, Rahim Entezari, Jonas Müller, Harry Saini, Yam Levi, Dominik Lorenz, Axel Sauer, Frederic Boesel, Dustin Podell, Tim Dockhorn, Zion English, Kyle Lacey, Alex Goodwin, Yannik Marek, and Robin Rombach.
\newblock Scaling rectified flow transformers for high-resolution image synthesis, 2024.
\newblock URL \url{https://arxiv.org/abs/2403.03206}.

\bibitem[Lin et~al.(2023)Lin, Yang, Li, Wang, and Wang]{lin2023designbenchexploringbenchmarkingdalle}
Kevin Lin, Zhengyuan Yang, Linjie Li, Jianfeng Wang, and Lijuan Wang.
\newblock Designbench: Exploring and benchmarking dall-e 3 for imagining visual design, 2023.
\newblock URL \url{https://arxiv.org/abs/2310.15144}.

\bibitem[Nichol et~al.(2022)Nichol, Dhariwal, Ramesh, Shyam, Mishkin, McGrew, Sutskever, and Chen]{nichol2022glidephotorealisticimagegeneration}
Alex Nichol, Prafulla Dhariwal, Aditya Ramesh, Pranav Shyam, Pamela Mishkin, Bob McGrew, Ilya Sutskever, and Mark Chen.
\newblock Glide: Towards photorealistic image generation and editing with text-guided diffusion models, 2022.
\newblock URL \url{https://arxiv.org/abs/2112.10741}.

\bibitem[Rombach et~al.(2022)Rombach, Blattmann, Lorenz, Esser, and Ommer]{rombach2022highresolutionimagesynthesislatent}
Robin Rombach, Andreas Blattmann, Dominik Lorenz, Patrick Esser, and Björn Ommer.
\newblock High-resolution image synthesis with latent diffusion models, 2022.
\newblock URL \url{https://arxiv.org/abs/2112.10752}.

\bibitem[Rapidata(2024)]{Rapidata}
Rapidata.
\newblock Rapidata: Human feedback on demand, 2024.
\newblock URL \url{https://www.rapidata.ai/}.

\bibitem[Alhabeeb and Al-Shargabi(2024)]{10431766}
Sarah~K. Alhabeeb and Amal~A. Al-Shargabi.
\newblock Text-to-image synthesis with generative models: Methods, datasets, performance metrics, challenges, and future direction.
\newblock \emph{IEEE Access}, 12:\penalty0 24412--24427, 2024.
\newblock \doi{10.1109/ACCESS.2024.3365043}.

\bibitem[Saharia et~al.(2022)Saharia, Chan, Saxena, Li, Whang, Denton, Ghasemipour, Ayan, Mahdavi, Lopes, Salimans, Ho, Fleet, and Norouzi]{saharia2022photorealistictexttoimagediffusionmodels}
Chitwan Saharia, William Chan, Saurabh Saxena, Lala Li, Jay Whang, Emily Denton, Seyed Kamyar~Seyed Ghasemipour, Burcu~Karagol Ayan, S.~Sara Mahdavi, Rapha~Gontijo Lopes, Tim Salimans, Jonathan Ho, David~J Fleet, and Mohammad Norouzi.
\newblock Photorealistic text-to-image diffusion models with deep language understanding, 2022.
\newblock URL \url{https://arxiv.org/abs/2205.11487}.

\bibitem[Ku et~al.(2024)Ku, Li, Zhang, Lu, Fu, Zhuang, and Chen]{ku2024imagenhubstandardizingevaluationconditional}
Max Ku, Tianle Li, Kai Zhang, Yujie Lu, Xingyu Fu, Wenwen Zhuang, and Wenhu Chen.
\newblock Imagenhub: Standardizing the evaluation of conditional image generation models, 2024.
\newblock URL \url{https://arxiv.org/abs/2310.01596}.

\bibitem[Wang et~al.(2023)Wang, Montoya, Munechika, Yang, Hoover, and Chau]{wang2023diffusiondblargescalepromptgallery}
Zijie~J. Wang, Evan Montoya, David Munechika, Haoyang Yang, Benjamin Hoover, and Duen~Horng Chau.
\newblock Diffusiondb: A large-scale prompt gallery dataset for text-to-image generative models, 2023.
\newblock URL \url{https://arxiv.org/abs/2210.14896}.

\bibitem[Feng et~al.(2023)Feng, He, Fu, Jampani, Akula, Narayana, Basu, Wang, and Wang]{feng2023trainingfreestructureddiffusionguidance}
Weixi Feng, Xuehai He, Tsu-Jui Fu, Varun Jampani, Arjun Akula, Pradyumna Narayana, Sugato Basu, Xin~Eric Wang, and William~Yang Wang.
\newblock Training-free structured diffusion guidance for compositional text-to-image synthesis, 2023.
\newblock URL \url{https://arxiv.org/abs/2212.05032}.

\bibitem[Bakr et~al.(2023)Bakr, Sun, Shen, Khan, Li, and Elhoseiny]{bakr2023hrsbenchholisticreliablescalable}
Eslam~Mohamed Bakr, Pengzhan Sun, Xiaoqian Shen, Faizan~Farooq Khan, Li~Erran Li, and Mohamed Elhoseiny.
\newblock Hrs-bench: Holistic, reliable and scalable benchmark for text-to-image models, 2023.
\newblock URL \url{https://arxiv.org/abs/2304.05390}.

\bibitem[Huang et~al.(2023)Huang, Sun, Xie, Li, and Liu]{huang2023t2icompbenchcomprehensivebenchmarkopenworld}
Kaiyi Huang, Kaiyue Sun, Enze Xie, Zhenguo Li, and Xihui Liu.
\newblock T2i-compbench: A comprehensive benchmark for open-world compositional text-to-image generation, 2023.
\newblock URL \url{https://arxiv.org/abs/2307.06350}.

\bibitem[Betker(2023)]{dalle-3-eval}
James Betker.
\newblock Dall-e 3 evaluation samples, 2023.
\newblock URL \url{https://github.com/openai/dalle3-eval-samples}.

\bibitem[Newman(2023)]{JMLR:v24:22-1086}
M.~E.~J. Newman.
\newblock Efficient computation of rankings from pairwise comparisons.
\newblock \emph{Journal of Machine Learning Research}, 24\penalty0 (238):\penalty0 1--25, 2023.
\newblock URL \url{http://jmlr.org/papers/v24/22-1086.html}.

\bibitem[Ouyang et~al.(2022)Ouyang, Wu, Jiang, Almeida, Wainwright, Mishkin, Zhang, Agarwal, Slama, Ray, Schulman, Hilton, Kelton, Miller, Simens, Askell, Welinder, Christiano, Leike, and Lowe]{ouyang2022traininglanguagemodelsfollow}
Long Ouyang, Jeff Wu, Xu~Jiang, Diogo Almeida, Carroll~L. Wainwright, Pamela Mishkin, Chong Zhang, Sandhini Agarwal, Katarina Slama, Alex Ray, John Schulman, Jacob Hilton, Fraser Kelton, Luke Miller, Maddie Simens, Amanda Askell, Peter Welinder, Paul Christiano, Jan Leike, and Ryan Lowe.
\newblock Training language models to follow instructions with human feedback, 2022.
\newblock URL \url{https://arxiv.org/abs/2203.02155}.

\end{thebibliography}






\end{multicols}
\end{document}